\documentclass[sigconf]{acmart}

\usepackage{graphicx}

\usepackage{xspace}
\usepackage[english]{babel}

\usepackage[utf8]{inputenc}
\usepackage[T1]{fontenc}
\usepackage{booktabs}
\usepackage{nicefrac}
\usepackage{xcolor}
\usepackage{enumitem}
\usepackage{lipsum}
\usepackage{soul}
\usepackage{threeparttable}
\usepackage{amsmath}
\usepackage{amsthm}

\usepackage{amssymb}
\usepackage{bm}
\usepackage{color}
\usepackage[ruled,linesnumbered]{algorithm2e}
\usepackage{mathtools}
\usepackage{multirow}
\usepackage{pifont}
\usepackage{pstricks-add}
\usepackage{subcaption}

\SetCommentSty{mycommfont}

\theoremstyle{definition}
\newtheorem{definition}{Definition}[]

\theoremstyle{definition}
\newtheorem*{remark}{Remark}

\copyrightyear{2024}
\acmYear{2024}
\setcopyright{acmlicensed}\acmConference[KDD '24]{Proceedings of the 30th ACM SIGKDD Conference on Knowledge Discovery and Data Mining}{August 25--29, 2024}{Barcelona, Spain}
\acmBooktitle{Proceedings of the 30th ACM SIGKDD Conference on Knowledge Discovery and Data Mining (KDD '24), August 25--29, 2024, Barcelona, Spain}
\acmDOI{10.1145/3637528.3672031}
\acmISBN{979-8-4007-0490-1/24/08}
\newcommand{\parh}[1]{\smallskip\noindent\textbf{#1}}

\newcommand{\highlight}[1]{\textcolor{orange}{#1}}
\newcommand{\improve}[1]{\scriptsize{\highlight{#1\%}}}

\newcommand{\spot}{\textsc{SPOT}\xspace}
\newcommand{\F}{Fig.}
\newcommand{\E}{Eqn.}

\newcommand{\T}{Table}
\renewcommand{\S}{Sec.}
\newcommand{\A}{Alg.}

\newcommand{\ignore}[1]{}

\def\sep{{\Perp_{\mathcal{G}}}}

\author{Pingchuan Ma}
\email{pmaab@cse.ust.hk}
\affiliation{
  \institution{HKUST}
  \country{Hong Kong SAR}
}

\author{Rui Ding}
\authornote{Corresponding author.}
\email{juding@microsoft.com}
\affiliation{
  \institution{Microsoft Research}
  \country{Beijing, China}
}

\author{Qiang Fu}
\email{qifu@microsoft.com}
\affiliation{
  \institution{Microsoft Research}
  \country{Beijing, China}
}

\author{Jiaru Zhang}
\email{jiaruzhang@sjtu.edu.cn}
\affiliation{
  \institution{Shanghai Jiao Tong University}
  \country{Shanghai, China}
}

\author{Shuai Wang}
\email{shuaiw@cse.ust.hk}
\affiliation{
  \institution{HKUST}
  \country{Hong Kong SAR}
}

\author{Shi Han}
\email{shihan@microsoft.com}
\affiliation{
  \institution{Microsoft Research}
  \country{Beijing, China}
}

\author{Dongmei Zhang}
\email{dongmeiz@microsoft.com}
\affiliation{
  \institution{Microsoft Research}
  \country{Beijing, China}
}


\begin{document}
\title{Scalable Differentiable Causal Discovery in the Presence of Latent
Confounders with Skeleton Posterior (Extended Version)}

\begin{abstract}
Differentiable causal discovery has made significant advancements in the
learning of directed acyclic graphs. However, its application to real-world
datasets remains restricted due to the ubiquity of latent confounders and the
requirement to learn maximal ancestral graphs (MAGs). To date, existing
differentiable MAG learning algorithms have been limited to small datasets and
failed to scale to larger ones (e.g., with more than 50 variables).

The key insight in this paper is that the causal skeleton, which is the
undirected version of the causal graph, has potential for improving accuracy and
reducing the search space of the optimization procedure, thereby enhancing the
performance of differentiable causal discovery. Therefore, we seek to address a
two-fold challenge to harness the potential of the causal skeleton for
differentiable causal discovery in the presence of latent confounders: (1)
scalable and accurate estimation of skeleton and (2) universal integration of
skeleton estimation with differentiable causal discovery.

To this end, we propose \spot (Skeleton Posterior-guided OpTimization), a
two-phase framework that harnesses skeleton posterior for differentiable causal
discovery in the presence of latent confounders. On the contrary to a
``point-estimation'', \spot seeks to estimate the posterior distribution of
skeletons given the dataset. It first formulates the posterior inference as an
instance of amortized inference problem and concretizes it with a supervised
causal learning (SCL)-enabled solution to estimate the skeleton posterior. To
incorporate the skeleton posterior with differentiable causal discovery, \spot
then features a skeleton posterior-guided stochastic optimization procedure to
guide the optimization of MAGs. 

Extensive experiments on various datasets show that \spot substantially
outperforms SOTA methods for MAG learning. \spot also demonstrates its
effectiveness in the accuracy of skeleton posterior estimation in comparison
with non-parametric bootstrap-based, or more recently, variational
inference-based methods. Finally, we observe that the adoption of skeleton
posterior exhibits strong promise in various causal discovery tasks.

\end{abstract}

\maketitle

\section{Introduction}

Causal discovery in the presence of latent confounders is a long-standing
problem~\cite{spirtes2000causation, zhang2008completeness}. Under this setting,
the causal relations are typically represented by a maximal ancestral graph
(MAG) \cite{spirtes2000causation}, a special class of acyclic directed mixed
graphs (ADMGs). MAG learning has historically been conducted by either
constraint-based methods, such as FCI \cite{spirtes2000causation,
zhang2008completeness}, RFCI \cite{colombo2012learning} and ICD
\cite{rohekar2021iterative}, or by score-based methods, such as M3HC
\cite{tsirlis2018scoring}, AGIP \cite{chen2021integer} and GPS
\cite{claassen2022greedy}. In recent years, differentiable causal discovery has
emerged as a promising approach to enhance the accuracy and efficiency of
existing methods~\cite{zheng2018dags,vowels2022d}. By recasting the
combinatorial constraint of graph structure into a differentiable form,
continuous optimization techniques can be applied in an ``out-of-the-box''
manner.

The goal of differentiable methods is to identify the ancestral ADMG, and apply
maximal ancestral projection, a standard process, to generate the corresponding
MAG. Despite the encouraging results achieved thus far, they struggle with
large-scale causal graphs, particularly those containing more than 20 variables.
One widely-used method, ABIC~\cite{bhattacharya2021differentiable}, acknowledges
its limitations with datasets of only 10--15 variables (i.e., a causal graph
with 10-15 nodes), which restricts its broader applicability. Likewise,
N-ADMG~\cite{ashman2022causal}, another differentiable method for ADMG learning,
suffers from similar scalability issues. Upon closer examination, we find that
these methods are both inefficient and inaccurate when working with large-scale
causal graphs. For example, ABIC may take days to converge and sometimes
produces cyclic graphs when dealing with 50 nodes. This is likely due to the
inherent challenges of learning from causally insufficient data, as ABIC must
manage a more complicated objective function, additional optimization variables,
and complex constraints. Consequently, the search space of large datasets poses
significant challenges for standard optimization techniques, such as
L-BFGS~\cite{fletcher1987practical} or gradient descent.

On the other hand, it has been demonstrated that adjacency information (i.e.,
the skeleton) can be utilized as a pre/post-processing step to facilitate
differentiable methods for learning DAGs (directed acyclic graphs). For
instance, CDRL~\cite{zhu2019causal} uses the CAM~\cite{buhlmann2014cam} to rule
out spurious edges while ML4S~\cite{ma2022ml4s} investigates using skeleton in a
pre-processing step to preclude superfluous variables for
NOTEARS~\cite{zheng2018dags}. These approaches show the potential for boosting
differentiable DAG learning by leveraging skeleton information.

\parh{Key Observation.}~These promising results on DAG learning motivate us to
investigate \textit{whether MAG learning can benefit from a high-quality
estimation of the skeleton}. To this end, we implement two ABIC variants, one
using the ground-truth skeleton as prior knowledge and the other using the
skeleton learnt by FCI, to validate our hypotheses. In both variants, we black
out the edges that are not in the skeleton and then run ABIC to only optimize
the remaining edges. In this way, the optimization procedure is only applied on
a subset of variables rather than the entire adjacency matrix. Then, we evaluate
them on ten synthetic datasets with 50-100 variables and 1000 samples (see
\T~\ref{tab:motivating-cmp}). In particular, we find that ABIC is impressively
accurate when the true skeleton is known (3rd row in
\T~\ref{tab:motivating-cmp}). This result, to a considerable extent, validates
our hypothesis that the skeleton can be used to reduce the search space of the
optimization procedure and therefore boost the performance of differentiable MAG
learning.

On the contrary, we also find that when the ground-truth skeleton is unknown and
the used skeleton is learned by FCI (4th row in \T~\ref{tab:motivating-cmp}),
the improvement becomes modest because many erroneous/missing edges induced by
FCI propagate to the subsequent optimization procedure. This indicates the
practical hurdle of using the skeleton to facilitate differentiable causal
discovery.

Indeed, due to finite samples, learning a high-quality skeleton is challenging.
Simply putting a ``\textit{point estimation}'' of the skeleton as a hard
constraint on the optimization procedure can result in considerable missing
edges (i.e., false negatives) and finally impair performance.

\begin{table}
  
  \centering
  \caption{Boosting differentiable causal discovery with skeleton information. 
  (avg. on ten datasets with 50-100 variables.)}
  \vspace{-10pt}
   \resizebox{0.85\linewidth}{!}{
  \begin{tabular}{l|l|l|l}
    \hline
    \textbf{Method} & \textbf{Skeleton F1} & \textbf{Arrowhead F1}& \textbf{Tail F1}\\
    \hline
    ABIC  & 0.84 & 0.76 & 0.67 \\\hline
    ~w/ True Skeleton  & 1.00 \improve{+19} & 0.96 \improve{+26}& 0.96 \improve{+33}\\\hline
    ~w/ FCI  & 0.87 \improve{+4}& 0.84 \improve{+11}& 0.71 \improve{+6}\\\hline
    \spot  & 0.91 \improve{+8}& 0.86 \improve{+13}& 0.78 \improve{+16}\\\hline

  \end{tabular}
   }
   \vspace{-10pt}
\label{tab:motivating-cmp}
\end{table}

The above preliminary results shed light on a possible solution that
synergistically combines differentiable causal discovery with \textit{skeleton}
information, while alleviating error propagations. Recently, much research has
promoted the value of posterior distribution of DAGs~\cite{cundy2021bcd,
lorch2021dibs, lorch2022amortized}. Enlightened by these works, we advocate
estimating a \textit{posterior distribution} (i.e., $p(S\mid \mathcal{D})$ where
$S$ is the skeleton and $\mathcal{D}$ is the dataset) over all skeletons to
replace the conventional ``point estimation.'' (i.e., $\arg\max_S p(S\mid
\mathcal{D})$ ) The posterior effectively quantifies \textit{epistemic
uncertainty and the degree of confidence} to any skeletons w.r.t.~the given
dataset. Nonetheless, unlike DAGs, skeletons over causally insufficient
variables do not have an explicit form of likelihood function (i.e.,
$p(\mathcal{D}, S)$). 

To address these challenges, we propose \spot\ (Skeleton Posterior-guided
OpTimization) as a two-phase framework for facilitating differentiable causal
discovery in the presence of latent confounders. \spot\ first performs amortized
variational inference to estimate skeleton posterior, \spot\ then employs a
novel optimization procedure for boosting subsequent differentiable causal
discovery. Specifically, \spot\ leverages a recent advancement to amortize the
variational inference into the joint distribution of data and corresponding
skeleton $p(\mathcal{D}, S)$ and alleviates the need of an explicit likelihood
function. Then, it concretizes the amortized inference with a supervised model
to estimate the skeleton posterior. To effectively facilitate the optimization
procedure, \spot\ employs the skeleton posterior to \textit{stochastically}
update variables in each optimization step, instead of {deterministically}
updating optimization variables with gradients.

As shown in the last row of \T~\ref{tab:motivating-cmp}, \spot\ improves the
performance of differentiable causal discovery methods by a notable margin. In
\S~\ref{sec:eval}, we conduct extensive experiments on various large-scale
datasets and show that \spot delivers 8\% improvement on skeleton F1 score and
13\% and 16\% improvement on arrowhead and tail F1 scores, respectively, which
are representative metrics for evaluating the accuracy of the learned ADMGs. We
also demonstrate that the skeleton posterior estimated by \spot is highly
accurate compared to other variational inference-based and non-parametric
bootstrap-based solutions. Finally, we explore the versatile applications of
\spot in MAG learning for non-linear causally insufficient datasets and also in
DAG learning methods. Our empirical results indicate the strong potential of
\spot in such scenarios.

In summary, we make the following contributions:

\begin{enumerate}[leftmargin=*,topsep=0pt,itemsep=0pt]
  \item Conceptually, we advocate a novel focus of using the skeleton posterior
  to facilitate differentiable causal discovery in the presence of latent
  confounders.
  \item Technically, we formulate the problem of skeleton posterior estimation
  under amortized inference framework and propose a supervised learning-based
  solution to estimate the skeleton posterior from observational data.
  \item On the basis of the skeleton posterior, we propose \spot, a novel
  stochastic optimization procedure, to facilitate differentiable causal
  discovery in the presence of latent confounders, which incorporates the
  skeleton posterior in a stochastic manner.
  \item Empirically, \spot\ demonstrates superior performance on nearly all
  evaluation metrics and various datasets, substantially improving upon its
  counterpart. We also explore the extension of \spot\ to other causal discovery
  settings, including non-linear ADMG and DAG. Our results show that \spot\
  consistently improves state-of-the-art methods in these settings. Our code
  will be made publicly available at~\cite{code}.
\end{enumerate}

\section{Preliminary}

As in many previous works~\cite{tsirlis2018scoring, chen2021integer,
claassen2022greedy, bhattacharya2021differentiable}, we focus on discovering MAG
from a linear Gaussian structural causal model (SCM) in the presence of latent
confounders and assume the absence of selection bias (no undirected edges in
ADMGs). In this section, we introduce preliminary knowledge of linear Gaussian
SCM with latent confounders, differentiable causal discovery and amortized
inference. In Appendix, we provide definitions to terminologies that are not
explicitly defined in the main text.

\subsection{Linear Gaussian SCM with Latent Confounders}
\label{subsec:linear-scm}

We first start with the definition of a linear Gaussian SCM without latent
confounders. Consider a linear SCM with $d$ observable variables parameterized
by a coefficient matrix $\delta\in\mathbb{R}^{d\times d}$. The SCM can be
written as
\begin{equation}
    \small
    \label{eq:linear-scm}
    V_i\leftarrow \sum_{V_j\in\textbf{PA}_i}\delta_{ji}V_j+\epsilon_i
\end{equation}
where $\textbf{PA}_i$ are the parents of $V_i$ and $\epsilon_i$ is a noise term
that is mutually independent of all other noise terms.

The noise term $\epsilon_i$ is mutually independent of others if and only if
$V_i$ has no latent confounder. Otherwise, $\epsilon_i$ is correlated with
$\epsilon_j$ if $V_j$ shares a latent confounder with $V_i$. When the noise
terms are Gaussian, the correlation can be expressed by a covariance matrix
$\beta=\mathbb{E}[\epsilon\epsilon^T]$ and the joint distribution marginalized
over observable variables $\bm{V}_\text{O}$ forms a zero-mean multivariate
Gaussian distribution with covariance matrix as
$\Sigma=(I-\delta)^{-T}\beta(I-\delta)^{-1}$. The induced graph ${G}$ is an ADMG
and contains two types of edges, including directed edges ($\rightarrow$) and
bidirected edges ($\leftrightarrow$), which implies two adjacency matrices,
namely $D$ and $B$. In the special case where there is no latent confounder, the
ADMG is equivalent to a DAG and the adjacency matrix of bidirected edges $B$ is
all zeros. $V_i\to V_j$ exists in ${G}$ and $D_{ij}=1$ if and only if
$\delta_{ij}\neq 0$. $V_i\leftrightarrow V_j$ exists in ${G}$ and $B_{ij}=1$ if
and only if $\beta_{ij}\neq 0$. It is commonly assumed that
$\delta_{ij}=\delta_{ji}=\beta_{ij}=\beta_{ji}=0$ if and only if
$D_{ij}=D_{ji}=B_{ij}=B_{ji}=0$. 

\begin{remark}
    \it
    In the context of ADMGs, the noise term $\epsilon$ does not exclusively
    represent the exogenous variables. In particular, $\epsilon_i$ in
    \E~\ref{eq:linear-scm} includes both the exogenous variable and the confounding
    effect from $V_u$ (the latent confounder). Similarly, the noise term on another
    variable $V_j$, $\epsilon_j$, also includes its own exogenous variable and the
    confounding effect from $V_u$. Therefore, $\epsilon_i$ and $\epsilon_j$ are
    jointly influenced by the confounding effect from $V_u$, which makes them
    dependent. We provide a more detailed explanation below.
    
    Consider the causal graph $V_{u_i} \rightarrow V_i \leftarrow V_u
    \rightarrow V_j \leftarrow V_{u_j}$. Without loss of generality on the
    linear setting, we have $V_i = V_{u_i} + aV_u$ and $V_j = V_{u_j} + bV_u$,
    where $a, b$ are non-zero coefficients. Given that only $V_i$ and $V_j$ are
    observed, the resulting ADMG should be $V_i \leftrightarrow V_j$, based on
    Sec. 2.2 of~\cite{zhang2008completeness}. In \E~\ref{eq:linear-scm}, $V_i =
    \sum_{V_k \in Pa(V_i)} \delta_{ki} V_k + \epsilon_i$ where $Pa(V_i)$ are the
    parents of $V_i$ in the ADMG. Since there is a bidirectional edge $V_i
    \leftrightarrow V_j$, neither $V_i$ is the parent of $V_j$ nor $V_j$ the
    parent of $V_i$. Hence, $Pa(V_i)=Pa(V_j)=\emptyset$. Thus, we have $V_i =
    \epsilon_i$ and $V_j = \epsilon_j$. However, since $V_i$ and $V_j$ are
    dependent, it follows that $\epsilon_i$ and $\epsilon_j$ must also be
    dependent.
\end{remark}

According to the causal Markov and faithfulness assumption, ancestral ADMG, MAG
and skeleton can be defined as follows.

\begin{definition}[Ancestral Acyclic Directed Mixed Graph]
    ${G}$ is an ancestral ADMG if it is a mixed graph with directed edges
    ($\rightarrow$) and bidirected edges ($\leftrightarrow$) and contains no
    directed cycles or almost directed cycles~\cite{zhang2008completeness}.
\end{definition}

\begin{definition}[Maximal Ancestral Graph]
    An ancestral ADMG $G$ is a Maximal Ancestral Graph if for each pair of
    non-adjacent nodes, there exists a set of nodes that make them
    m-separated~\cite{zhang2008completeness}.
\end{definition}

\begin{definition}[Skeleton]
    An undirected graph $S$ is a skeleton of an MAG $G$ if $S$ is obtained from
    $G$ by replacing all directed edges with undirected edges.
\end{definition}

Therefore, two nodes $V_i,V_j$ are adjacent in the skeleton if and only
if~$\forall \bm{Z}\subseteq \bm{V}_\text{O}\setminus \{V_i,V_j\}, V_i \not\perp
V_j\mid \bm{Z}$, where $\perp$ and $\not\perp$ denote conditional independence
and dependence, respectively. Hence, it is clear that given a dataset whether
two nodes are adjacent is not influenced by the adjacencies of other nodes.

\parh{Parameter Estimation of Linear Gaussian SCM.}~For DAGs, one easily can
estimate the parameters of the SCM by solving a simple least squares regression
problem. However, for ADMGs, the estimation of the parameters is more
challenging due to the presence of latent confounders.
\citet{drton2009computing} proposed an iterative algorithm to estimate the
parameters of the SCM called Residual Iterative Conditional Fitting (RICF). RICF
generally works for bow-free ADMGs and ancestral ADMGs, which are our focus in
this paper, are special cases of bow-free
ADMGs~\cite{bhattacharya2021differentiable}. This algorithm iteratively fits the
SCM to the data and updates the covariance matrix. The algorithm is guaranteed
to converge to a local minimum when the corresponding ADMGs are ancestral.

\subsection{Differentiable Causal Discovery}

Score-based methods aim to maximize the score (e.g., log-likelihood) of the
graph on the given data, which can be written in the following form:
\begin{equation}
    \small
    \arg\max_{{G}} f(G) \text{ s.t. } {G} \text{~is acyclic}
\end{equation}
where $f(\cdot)$ is the score function. Given the acyclicity constraint, the
optimization process is combinatorial. Recently, differentiable methods
reformulate this combinatorial constraint into a constraint
$h_{\text{DAG}}(W)=\text{tr}(e^{W\circ W})-d$ such that
\begin{equation}
    \label{eq:dag-constraint}
    \small
    h_{\text{DAG}}(W)=0\iff {G} \text{~is acyclic}
\end{equation}
\noindent where $d=|\bm{V}_\text{O}|$ and $W$ is a weighted adjacency matrix of
${G}$. For linear SCM, an element in $W$ represents the linear coefficient. As a
continuous optimization with equality constraints, the augmented Lagrangian
method (ALM) is commonly used to convert the constrained optimization problem
into several unconstrained subproblems and use standard optimizers to solve them
separately~\cite{zheng2018dags}.

Despite the success of differentiable methods for DAG learning, the algebraic
characterization in \E~\ref{eq:dag-constraint} cannot be directly applied to
ADMGs (and MAGs). As aforementioned, ADMGs requires two adjacency matrices, $D$
and $B$, to represent directed edges and bidirected edges, respectively. To
extend the algebraic characterization to ADMGs,
ABIC~\cite{bhattacharya2021differentiable} modified it as 
\begin{equation}
    \label{eq:admg-constraint}
    h_{\text{ADMG}}(D,B)=\text{tr}(e^{D})-d+\text{sum}(e^{D}\circ B)
\end{equation}
where $\circ$ denotes the Hadamard product, i.e., element-wise multiplication,
and $\text{sum}(\cdot)$ is the sum of all elements in a matrix. It has been
proved that $h_{\text{ADMG}}(D, B)=0\iff {G} \text{~is ancestral ADMG}$. In a
nutshell, $\text{tr}(e^{D})-d$ implies the standard directed acyclicity
constraint and the last term $\text{sum}(e^{D}\circ B)$ is a term to ensure that
the bidirected edges does not introduce ``almost directed cycles''. In the
following, we use $h$ to denote $h_{\text{ADMG}}$ for simplicity.

\begin{algorithm}[t]
\small
\caption{ABIC~\cite{bhattacharya2021differentiable}}
\label{alg:abic}
\KwIn{Dataset $\mathcal{D}$}
\KwOut{Adjacency Matrix $D,B$}
Initialize $\delta^{(1,1)},\beta^{(1,1)}$\;
Define $h(\delta,\beta)$ according to \E~\ref{eq:admg-constraint}\;
\ForEach{$t_1=1,\cdots,T_{\text{ALM}}$}{
    \ForEach{$t_2=1,\cdots,T_{\text{RICF}}$}{
        Update pseudovariables by $\delta^{(t_1,t_2)},\beta^{(t_1,t_2)}$\;
        Constitute $f(\delta,\beta)$\;
        $\delta^{(t_1,t_2+1)},\beta^{(t_1,t_2+1)}\leftarrow \arg\min f(\delta,\beta)$\;
    }
    Update $\alpha^{(t_1)},\rho^{(t_1)},\lambda^{(t_1)}$\;
}
$D\leftarrow \bm{1}(|\delta^{(T_{\text{ALM}},T_{\text{RICF}})} > \omega|)$\; 
$B\leftarrow \bm{1}(|\beta^{(T_{\text{ALM}},T_{\text{RICF}})}> \omega|)$\;
\Return $D,B$\;
\end{algorithm}

Now, we introduce ABIC~\cite{bhattacharya2021differentiable} in
\A~\ref{alg:abic}. To ease presentation, we omit technical details (e.g.,
calculations of $\alpha,\rho,\lambda$, pseudovariables, and stopping criteria)
and refer readers to the original paper for the full version. The ABIC algorithm
aims to maximize the likelihood of ADMG over the observational data
$\mathcal{D}$, ensuring $h_{\text{ADMG}}(\delta,\beta)=0$. It employs two nested
loops: the outer loop applies the augmented Lagrangian method (ALM) to convert
the equality-constrained problem into $T_{\text{ALM}}$ steps of unconstrained
ones (lines 3--10 in \A~\ref{alg:abic}). After ALM, $\delta$ and $\beta$
represent the ADMG's linear coefficients and covariance, respectively. The
adjacency matrix is derived by applying a threshold to remove negligible
coefficients (lines 22--23), yielding the ancestral ADMG. If a MAG is required,
the maximal ancestral projection, a standard post-processing procedure, is
performed.

Inside the inner loop, \A~\ref{alg:abic} iteratively adjust the objective
function (lines 5--6) and solve the unconstrained optimization problem
accordingly (line 7). In particular, the objective function $f(\delta,\beta)$
(line 6) consists of the least square loss to fit the ADMG to the data, the
constraint $h(\delta,\beta)$, and a regularization term to enforce sparsity. The
form of $f(\delta,\beta)$ is updated according to the current pseudovariables
(line 5). Then, in the optimization phase (line 7), optimization algorithms are
used in an ``out-of-the-box'' manner to find a minimum of $f(\delta,\beta)$.
However, due to the complexity and non-convex nature of $f(\delta,\beta)$, it is
challenging for standard optimization techniques to find a plausible solution,
especially when the ADMG is large, as we will show in \S~\ref{sec:eval}.

\begin{figure*}[!htbp]

    \centering
\includegraphics[width=0.8\linewidth]{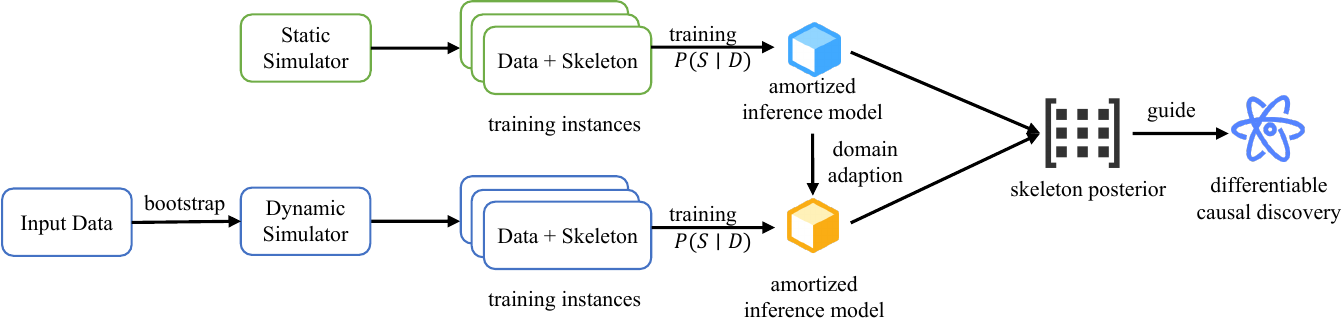}
\vspace{-5pt}
\caption{\spot\ workflow.}
\label{fig:workflow}

\end{figure*}

\subsection{Amortized Inference for Causal Discovery}

In the conventional variational inference framework, the posterior distribution
is approximated by a parametric distribution $q_\phi(G|\mathcal{D})$ with
parameters $\phi$. Here, $G$ is the causal graph and $\mathcal{D}$ is the
observed data in the context of causal discovery. The parameters are learned by
minimizing the KL divergence between the true posterior $p(G|\mathcal{D})$ and
the parametric distribution $q_\phi(G|\mathcal{D})$. Typically, variational
inference requires the likelihood function $p(\mathcal{D}, G)$ to compute the KL
divergence. Recently, amortized inference has been proposed to alleviate the
need of an explicit likelihood function by introducing a simulator that can
yield samples from the joint distribution $p(G,
\mathcal{D})$~\cite{ambrogioni2019forward}. \citet{lorch2022amortized} follows
this line of work and introduces AVICI, which amortizes the inference process by
introducing a sampler that can generate samples from the simulator and capture
the domain-specific inductive biases that would have been difficult to
characterize (e.g., gene regulatory networks or non-linear functions of random
Fourier features).

Recently, SCL (supervised causal learning) has emerged as a promising paradigm
for causal discovery~\cite{ke2022learning, ma2022ml4s,dai2023ml4c}. Compared to
standard machine learning models, SCL models features many advantages tailored
for causal discovery (e.g., invariant to the permutation of variables). It is
worth noting that the concept of ``permutation invariance'' is different from
the one in machine learning setting. One dataset (a $M\times N$ matrix) is a
sample to the SCL model. It implies that the model is permutation in- and
equivariant with respect to the observation and variable dimensions of the
provided dataset, respectively~\cite{ke2022learning}. SCL models can be trained
on samples from a known simulator. In essence, SCL can be viewed as an
instantiation of amortized inference when the simulator is in line with the
underlying causal mechanism of the observational data.
\section{\spot}

As aforementioned, skeleton information can be leveraged to enhance the
optimization procedure of differentiable causal discovery in the presence of
latent confounders. However, a ``point estimation'' of the skeleton is
prone-to-error. In this section, we propose \spot\ (Skeleton Posterior-guided
OpTimization) to leverage a probabilistic skeleton posterior to guide the
optimization of ADMGs.

As shown in \F~\ref{fig:workflow}, \spot consists of four steps: \ding{192} a
static simulator first generates an initial set of data/skeleton pairs to obtain
an initial amortized inference model (blue one in \F~\ref{fig:workflow}).
\ding{193} while the initial amortized inference model already provides a good
estimation of the skeleton posterior, \spot can optionally use a dynamic
simulator to generate more \textit{in-domain} training instances with respect to
the input dataset. \ding{194} with samples from the dynamic simulator, \spot
adapts the initial amortized inference model to obtain an updated model (yellow
one in \F~\ref{fig:workflow}) for skeleton posterior inference. \ding{195} \spot
uses the inferred skeleton posterior to enhance the optimization procedure of
differentiable causal discovery.

\parh{Conceptual Complexity.}~While our solution provides an additional layer of
complexity compared to standard differentiable causal discovery, we argue that
the level of complexity is generally manageable and comparable to other causal
discovery algorithms. First, the two stages in our pipeline are independent and
do not involve joint training. Second, constraint-based causal discovery
algorithms, such as FCI, also involve two similar steps: learning the skeleton
in the first step and orienting the edges in the second step.

\subsection{Skeleton Variational Objective}
\label{subsec:variational}
Let $\mathcal{D}=\{\bm{x}^1,\cdots,\bm{x}^n\}\sim p(V)$ be the observational
dataset, where $\bm{x}^i$ is sampled from the joint distribution $p(V)$ and $n$
is sample size. We aim to approximate the posterior over skeletons $p(S\mid
\mathcal{D})$ with a variational distribution $q(S;a)$ where $S$ is the
(symmetric) adjacency matrix of the skeleton and $a$ is the variational
parameters. Thus, the skeleton estimation can be decomposed into a set of edge
estimations (i.e., the probability of adjacency) independently. Then, the
variational family of $q(S;a)$ is defined as 
\begin{equation}
    \small
    q(S;a)=\prod_{i<j}q(S_{ij};a_{ij}) \text{ with } S_{ij}\sim \text{Bernoulli}(a_{ij}).
\end{equation}

In this regard, we aim to find an inference model $f_\phi(\mathcal{D})$ that
predicts $a$. This procedure can be attained by minimizing the expected forward
KL divergence from $p(S\mid \mathcal{D})$:

\begin{equation}
    \small
    \label{eq:kl}
    \min_{\phi}\mathbb{E}_{p(\mathcal{D})} \left[\text{KL}\left(p(S\mid \mathcal{D})\mid\mid q(S;f_\phi(\mathcal{D}))\right)\right]
\end{equation}

Following the principle of amortized inference~\cite{ambrogioni2019forward}, we
amortize the inference and rewrite the objective as 
\begin{equation}
    \small
    \begin{aligned}
        &\mathbb{E}_{p(\mathcal{D})} \text{KL}\left(p(S\mid \mathcal{D})\mid\mid q(S;f_\phi(\mathcal{D}))\right)\\
        =& \mathbb{E}_{p(\mathcal{D})} \mathbb{E}_{p(S\mid \mathcal{D})} \left[\log p(S\mid\mathcal{D}) - \log q(S;f_\phi(\mathcal{D}))\right]\\
        =& - \mathbb{E}_{p(S)} \mathbb{E}_{p(\mathcal{D}\mid S)}\left[\log q(S;f_\phi(\mathcal{D}))\right] + \text{ const.}\\
        =& - \mathbb{E}_{p(\mathcal{D}, S)}\left[\log q(S;f_\phi(\mathcal{D}))\right] + \text{ const.}
    \end{aligned}
\end{equation}
Since the constant does not depend on $\phi$, we can merely minimize
$\mathcal{L}(\phi)\coloneqq -\mathbb{E}_{p(\mathcal{D}, S)}\left[\log
q(S;f_\phi(\mathcal{D}))\right]$ to obtain $\phi$. In other words, the problem
is recast to train a predictive model $f_{\phi}(\cdot):\mathcal{D}\mapsto S$
over the distribution $p(\mathcal{D}, S)$.

\parh{Estimate $p(\mathcal{D}, S)$ via Static Simulator.}~Naturally, we can
estimate $p(\mathcal{D},S)$ by using a simulator (e.g., Erdős-Rényi random graph
model) then generate the corresponding dataset with pre-defined functional
forms. The simulator serves a direct sampler of $p(\mathcal{D}, S)$ if the test
data $\mathcal{D}$ (i.e., the input dataset on which we need to conduct causal
discovery) is in the same distribution of the \textit{static simulator}. In
other words, the pair of $\mathcal{D}$ and the skeleton $S$ is known to be drawn
from $p$ a priori. 

\parh{Estimate $p(\mathcal{D}, S)$ via Dynamic Simulator.}~To further feed the
model with more \textit{in-domain} training instances, we propose to use
nonparametric bootstrap method~\cite{friedman1999data} to estimate a MAG
$\hat{G}$ with a random subset of observational data. Then, we fit the
parameters of the underlying SCM (i.e., $\delta,\beta$; see
\S~\ref{subsec:linear-scm}) on the given test observational data using the RICF
algorithm~\cite{drton2009computing}. Finally, we regenerate the new
observational data $\hat{\mathcal{D}}$ from the fitted SCM. By repeating the
above procedure, we obtain a set of data/MAG pairs
$\{(\hat{\mathcal{D}}_1,\hat{G}_1),\cdots\}$ from which we can derive the
samples of $p(\mathcal{D}, S)$ as $\{(\hat{\mathcal{D}}_1,\hat{G}_1+
\hat{G}_1^T),\cdots\}$.\footnote{$G$ and $G^T$ are the adjacency matrix and its
transpose, $\hat{G}_1+ \hat{G}_1^T$ is the skeleton adjacency matrix.} We note
that, while the way latent variables affect the observed variables is implicit,
in the linear Gaussian setting, we can characterize the marginalized
distribution over observed variables as a zero-mean multivariate Gaussian
distribution with a covariance matrix defined as
$\Sigma=(I-\delta)^{-T}\beta(I-\delta)^{-1}$, where $\delta$ and $\beta$ are
parameters fitted by RICF. In this way, we presume this \textit{dynamic
(dataset-dependent) simulator} would provide more relevant training instances
with respect to the input dataset $\mathcal{D}$. 

\begin{remark}
    {\it In summary, the static simulator features a robust model that generalizes
    well to the test data, even with potential domain shift, as validated
    in~\cite{lorch2022amortized, ke2022learning}. On the other hand, the dynamic
    simulator provides more in-domain training instances, which may lead to a
    better performance at a fairly lightweight cost on the runtime, as will be
    shown shortly in \S~\ref{subsec:espi}.}
\end{remark}

\subsection{Skeleton Posterior Inference}
\label{subsec:espi}

\begin{figure}[!htbp]

    \centering
\includegraphics[width=\linewidth]{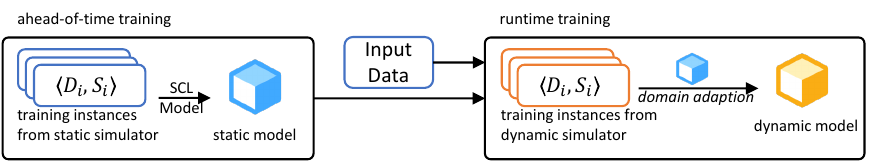}

\caption{Skeleton Inference Procedure.}
\label{fig:infer}

\end{figure}

In light of \S~\ref{subsec:variational}, the predictive model $f_{\phi}(\cdot):
\mathcal{D}\mapsto S$ can be instantiated by any supervised learning model.
Enlightened by the prosperous progress in SCL, we propose to use SCL model to
enable the amortized inference and estimate the skeleton posterior $p(S\mid
\mathcal{D})$. Below, we elaborate the design considerations of instantiating
the SCL model in \spot.

\ding{192} \textit{Model Architecture.}~Existing SCL methods either use an
end-to-end model (e.g., Transformer) to directly predict the adjacency
matrix~\cite{ke2022learning} or use a simple classifier (e.g., xgboost) to
predict the local structure (e.g., the adjacency of two variables or the
v-structure)~\cite{ma2022ml4s,dai2023ml4c}. While the design of \spot is
agnostic to the model architecture, we anticipate to use a simple classifier, as
it is more effective and efficient for estimating $p_{ij}$ in practice. Hence,
in our implementation, we adopt a ML4S-like cascade model~\cite{ma2022ml4s} to
predict the adjacency and constitute the skeleton posterior. We formulate the
problem of efficient skeleton posterior inference as predicting a series of
adjacencies and construct the skeleton posterior. Specifically, we aim to
estimate
\begin{equation}
    \small
    f_{\phi}(\mathcal{D})_{ij}=\mathbb{E}_{p(\mathcal{D}, S)}\left[p(V_i-V_j\in S\mid \mathcal{D})\right],
\end{equation}
We focus on the probability $p(V_i-V_j\mid\mathcal{D})$ for the adjacency of
$V_i,V_j$ given $\mathcal{D}$, which helps constitute the skeleton posterior (we
use $p_{ij}$ as an abbreviation of $p(V_i-V_j\mid\mathcal{D})$). Besides, we
also reproduce \citet{ke2022learning}'s Transformer-based model, and the results
show that the cascade model is more effective and efficient.

\ding{193} \textit{Objective Refinement.}~Instead of standard causal discovery
objectives (e.g., the higher F1 score), we prefer higher recall over precision
to prevent missed edges. This is because the spurious edges will be offloaded to
and eventually removed by differentiable causal discovery algorithms.
Specifically, even though the SCL model may predict a spurious edge and
corresponding parameters $\delta_{ij},\beta_{ij}$ tend to be non-zero, the
differentiable causal discovery algorithm will gradually move the coefficients
to zero to minimize the objective function. In this regard, the spurious edges
are eventually removed. However, the missed edges are hard to recover due to the
sparse nature of the ADMG. To this end, we can either set a conservative
threshold (subject to the specific design of the SCL model) or apply label
smoothing trick to the training instances to prevent the model from predicting
zero probability for the true edges.

\ding{194} \textit{Efficient Training.}~When using the dynamic simulator, the
model has to be trained from scratch with the training instances sampled from
non-parametric bootstrap. However, this would impose a heavy sampling and
training overheads in the runtime. To alleviate this issue, we adopt a
widely-used \textit{domain adaptation paradigm} to reduce the training overheads
by leveraging the static model. Specifically, given a dataset, we only sample a
few training instances from the dynamic simulator. Then, we use the output and
intermediate values of the static model to augment the original features of the
samples from the dynamic simulator and train an adapted dynamic model
efficiently. In this way, the dynamic model can rapidly adapt to the new dataset
with few-shot training instances and yields a better performance with mild cost.

\subsection{Posterior-guided Optimization}
\label{subsec:optim}

Non-adjacency between $V_i,V_j$ implies $\delta_{ij} = \delta_{ji} = \beta_{ij}
= \beta_{ji} = 0$. When ground-truth skeleton is available, we enforce extra
equality constraints to optimize. However, the true skeleton is unattainable in
practice, necessitating estimating a skeleton from observational data. We
advocate using the skeleton posterior for optimization as it encodes epistemic
uncertainty.

It is unclear how to incorporate a skeleton posterior in the optimization
procedure. The estimated skeleton posterior from \S~\ref{subsec:espi} is a
continuous adjacency matrix $p_{ij}\in[0,1]$. Thus, we cannot derive meaningful
constraints for coefficients/covariances. Formally, $p(V_i-V_j\mid
\mathcal{D})=p_{ij}$ implies the probability of the union of (disjoint) events
$P(\delta_{ij}\neq 0 \cup \delta_{ji}\neq 0 \cup \beta_{ij}\neq 0 \mid
\mathcal{D})$ equals $p_{ij}$. Higher posterior probability indicates higher
likelihood of non-zero coefficients. In this regard, $p_{ij}$ cannot provide any
additional information regarding the value of non-zero coefficients. For
example, $\delta_{ij}=0.1$ and $\delta_{ij}=0.9$ both comply with arbitrary
positive posterior $p_{ij}>0$ equally well. To account for the probabilistic
nature of the skeleton posterior, we propose the posterior-guided optimizer for
structure learning as a replacement of the standard optimizer in differentiable
causal discovery (e.g., the gradient descent or the L-BFGS). 

\begin{algorithm}[t]
\caption{Skeleton Posterior-guided Optimizer}
\label{alg:spot}
\KwIn{Skeleton Posterior $p$, Optimization Variables $\delta,\beta$, Objective Function $f$, ALM Step $t$, Temperature Constant $c$}
\KwOut{Optimized Variables $\delta,\beta$}
$\delta^{*},\beta^{*} \leftarrow \arg\min f(\delta,\beta)$\;
\ForAll{$\delta_{ij}$ \textbf{and} $\beta_{ij}$}{
    \eIf{$\delta_{ij} \times \frac{\partial f(\delta,\beta)}{\partial\delta_{ij}} > 0$}{
        $\delta_{ij} \leftarrow \delta_{ij}^{*}$
    }{
        Update $\delta_{ij} \leftarrow \delta_{ij}^{*}$ with probability $(p_{ij} + c)^{\frac{1}{t+1}}$\;
        
    }
}
\Return $\delta,\beta$\;
\end{algorithm}

\begin{figure}[h]
    
	\centering
	\includegraphics[width=0.7\linewidth]{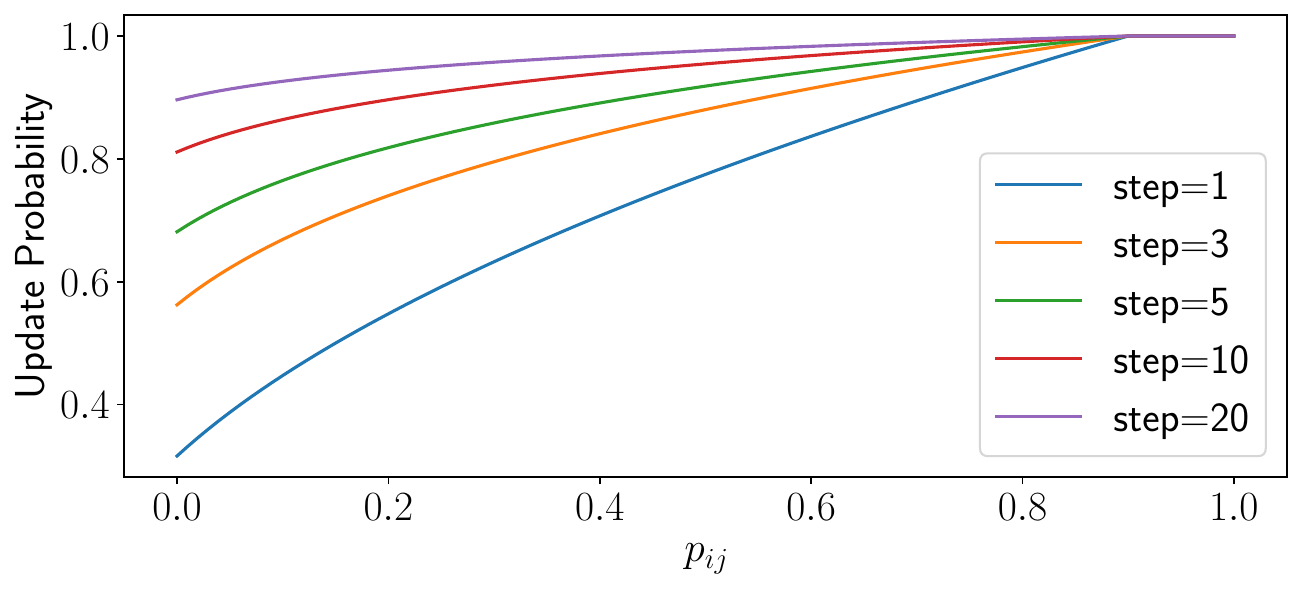}
    
	\caption{The probability of accepting an update for different $p_{ij}$ and
    $t$ with $c=0.1$.}
    
    \label{fig:prob-accept}
\end{figure}

\A~\ref{alg:spot} presents the skeleton posterior-guided optimizer. In
accordance with many standard differentiable causal discovery methods, we first
optimize the objective function $f(\delta,\beta)$ with the standard optimizer
(e.g., gradient descent) to obtain the optimal coefficients and covariances
$\delta^{*},\beta^{*}$ (line 1). Then, instead of directly taking
$\delta^{*},\beta^{*}$ as the final result, we introduce a stochastic update
scheme to update $\delta,\beta$ based on the skeleton posterior $p$ (lines 2--8)
as follows:
\begin{equation}
    \small
P(\delta_{ij}\leftarrow\delta_{ij}^*) = \begin{cases}
    1  & \delta_{ij} \times \frac{\partial f(\delta,\beta)}{\partial\delta_{ij}} >
    0 \\
    (p_{ij}+c)^{\frac{1}{t+1}} & \text{otherwise}
  \end{cases}
\end{equation}
If $\delta_{ij} \times \frac{\partial f(\delta,\beta)}{\partial\delta_{ij}} >
0$, the update is accepted unconditionally (line 3--4); otherwise, it is
accepted with a probability of $(p_{ij}+c)^{\frac{1}{t+1}}$ (line 12--15;
\F~\ref{fig:prob-accept}). If rejected, we keep $\delta_{ij}$ unchanged (line
13). $\beta$ is also optimized in a similar way.

Conceptually, the overall procedure can be viewed as a modified simulated
annealing process~\cite{kirkpatrick1983optimization}. $\delta_{ij} \times
\frac{\partial f(\delta,\beta)}{\partial\delta_{ij}} > 0$ implies that the
coefficient is ``moving to zero'' (i.e., a weakened edge) in this update. Given
that real-world ADMGs are often sparse, ``moving-to-zero'' update is encouraged
and therefore accepted unconditionally. This design consideration is thus an
analogy of the unconditional update in simulated annealing. 

In contrast, if the coefficient is ``moving to non-zero'' (i.e., a strengthened
edge), the update is accepted with a probability computed by $p_{ij}$ and $t$
jointly. Recall that $p_{ij}$ is the probability of the edge existence. The
update is thus more encouraged with higher $p_{ij}$. $\frac{1}{t}$ is a analogy
of the temperature parameter in simulated annealing. As shown in
\F~\ref{fig:prob-accept}, an update is more likely to be rejected with lower $t$
and higher $t$ yields $(p_{ij}+c)^{\frac{1}{t}}\approx 1$. With the increase of
$t$, the probability of accepting an update is gradually similar to what would
be in the vanilla optimization procedure even with a low $p_{ij}$. In this way,
the optimization procedure is gradually stabilized. Furthermore, it is worth
noting that the scheme is robust to the accuracy of the skeleton posterior. For
spurious edges, even if $p_{ij}$ is high, the optimization procedure of
differentiable causal discovery is still able to move the coefficients to zero
given the unconditional update for ``edge-weakening'' updates.

\begin{table*}[t]
	\caption{End-to-end comparison~(avg.~on ten datasets with nodes $\in [50, 100]$).}

	\centering
	\resizebox{\textwidth}{!}{
		\begin{tabular}{l|ccc|ccc|ccc|c}
			\toprule
			\multirow{2}{*}{\textbf{Method}} & \multicolumn{3}{c|}{\textbf{Skeleton}}  & \multicolumn{3}{c|}{\textbf{Arrowhead}} & \multicolumn{3}{c|}{\textbf{Tail}}  & \multicolumn{1}{c}{\textbf{\#Failed}}  \\ 
			\cmidrule(lr){2-4} \cmidrule(lr){5-7} \cmidrule(lr){8-10} 
			 & F1 & TPR & FDR & F1 & TPR & FDR & F1 & TPR & FDR & {\bf Datasets} \\ 
			\midrule
			FCI
			& 0.84 $\pm$ 0.06& 0.79 $\pm$ 0.10& 0.10 $\pm$ 0.05& 0.57 $\pm$ 0.05& 0.78 $\pm$ 0.15& 0.54 $\pm$ 0.04& 0.53 $\pm$ 0.18& 0.46 $\pm$ 0.20& 0.31 $\pm$ 0.15 & 0/10 \\
			RFCI
			& 0.85 $\pm$ 0.07& 0.77 $\pm$ 0.09& 0.05 $\pm$ 0.03& 0.62 $\pm$ 0.09& 0.76 $\pm$ 0.12& 0.45 $\pm$ 0.12& 0.51 $\pm$ 0.15& 0.42 $\pm$ 0.17& 0.27 $\pm$ 0.14& 0/10\\
			ICD 
			& 0.82 $\pm$ 0.09& 0.81 $\pm$ 0.07& 0.17 $\pm$ 0.11& 0.56 $\pm$ 0.10& 0.82 $\pm$ 0.06& 0.57 $\pm$ 0.12& 0.48 $\pm$ 0.15& 0.41 $\pm$ 0.15& 0.39 $\pm$ 0.18& 1/10\\
			M3HC
			& 0.73 $\pm$ 0.09& 0.59 $\pm$ 0.12 & {\bf 0.03}$\pm$ 0.03& 0.56 $\pm$ 0.15& 0.46 $\pm$ 0.16& 0.27 $\pm$ 0.11& 0.32 $\pm$ 0.11& 0.24 $\pm$ 0.10& 0.48 $\pm$ 0.13& 0/10\\
			GPS
			& 0.73 $\pm$ 0.10 & 0.81 $\pm$ 0.11& 0.33 $\pm$ 0.11& 0.62 $\pm$ 0.09& 0.85 $\pm$ 0.11 & 0.50 $\pm$ 0.10& 0.43 $\pm$ 0.12& 0.62 $\pm$ 0.19& 0.66 $\pm$ 0.10& 0/10\\
			ABIC
			& 0.84 $\pm$ 0.04 & 0.89 $\pm$ 0.07 & 0.19 $\pm$ 0.07& 0.76 $\pm$ 0.07& 0.85 $\pm$ 0.15& 0.29 $\pm$ 0.11& 0.67 $\pm$ 0.07& 0.80 $\pm$ 0.15& 0.40 $\pm$ 0.07& 0/10\\ \midrule
			\spot
			& {\bf 0.91} $\pm$ 0.03& {\bf 0.94} $\pm$ 0.04 & 0.11 $\pm$ 0.04 & {\bf 0.86} $\pm$ 0.05 & {\bf 0.89}  $\pm$ 0.10& {\bf 0.16}  $\pm$ 0.06& {\bf 0.78} $\pm$ 0.09 & {\bf 0.86} $\pm$ 0.13 & {\bf 0.27} $\pm$ 0.13 & 0/10 \\ 
			\bottomrule
		\end{tabular}
	} 
	
	\label{tab:overall-cmp}
\end{table*}

\parh{Convergence.}~In the context of differentiable causal discovery, the
concept of convergence is three-fold. First, in accordance with the
ABIC~\cite{bhattacharya2021differentiable}, asymptotically, convergence to the
global optimum of its objective function implies that the corresponding ADMG is
within the Markov equivalence class of the true ADMG. Second, given the
non-convexity of the objective function, it is not guaranteed that the
optimization procedure converges to the global optimum; instead, it may result
in a local minimum or a saddle point. Finally, we discuss the impact of the
stochastic update scheme on convergence to the local minima in the following
sense: Since the probability of an update reaches 1 (see
\F~\ref{fig:prob-accept}), the optimization procedure becomes equivalent to the
vanilla optimizer when $t$ is sufficiently large. In this regard, the stochastic
update scheme does not negatively affect the convergence to the local minima. In
fact, our empirical results show that our stochastic update scheme yields much
better results than ABIC and also surpasses other SOTA methods.

\parh{Probabilistic Update vs. Regularization.}~A natural question arises: why
not use regularization (e.g., a penalty term) to incorporate the skeleton
posterior? We argue that the penalty term may face several hurdles in practice.
First, the objective function is often unstable due to the acyclic term
(\E~\ref{eq:admg-constraint}). Its value can range from $1 \times 10^{-5}$ to $1
\times 10^5$ over different stages of the optimization. Second, as
aforementioned, $\delta,\beta$ are not directly comparable with posterior
probability. This necessitates considerable efforts for implementation, thus we
did not adopt it.

\section{Evaluation}
\label{sec:eval}
In evaluation, we aim to evaluate the performance of \spot from three
perspectives: \ding{202} end-to-end causal discovery performance; \ding{203}
standalone skeleton posterior inference performance; \ding{204} extension to
other differentiable causal structure learning algorithms. We also compare \spot
with a wide range of baselines.

\parh{Data Generation.}~We follow the similar setting from previous
works~\cite{bernstein2020ordering,bhattacharya2021differentiable} to generate
synthetic datasets. We generates ADMGs from Erdős-Rényi (ER) model with various
nodes whose average indegree is $\in [1,1.5]$. Each ADMG contains $5-15\%$
bidirected edges. Then, it is parameterized by the same rules of
\cite{bhattacharya2021differentiable}: if $V_i\to V_j$, $\delta_{ij}$ is
uniformly sampled from $\pm [0.5, 2.0]$; if $V_i\leftrightarrow V_j$,
$\beta_{ij}=\beta_{ji}$ is uniformly sampled from $\pm [0.4, 0.7]$; and
$\beta_{ii}$ is uniformly sampled from $\pm [0.7, 1.2]$ and add
$\text{sum}(|\beta_{i,-i}|)$ to ensure positive definiteness. For each ADMG, we
generate 1000 samples.

\parh{Baselines.}~We compare our method with a wide range of baselines,
including constraint-based methods (FCI~\cite{spirtes2000causation},
RFCI~\cite{colombo2012learning} and ICD~\cite{rohekar2021iterative}),
score-based methods (M3HC~\cite{tsirlis2018scoring} and
GPS~\cite{claassen2022greedy}) and also the differentiable causal discovery
methods (ABIC~\cite{bhattacharya2021differentiable}). The comparison is made on
synthetic datasets with 50--100 nodes. We observe that ABIC occasionally outputs
cyclic graphs with the default threshold, which is unwarranted for maximal
ancestral projection. In such cases, we use the minimal threshold that produces
an acyclic graph. We also tentatively tried AGIP~\cite{chen2021integer}
algorithm and excluded it from comparison, because its preprocessing step is
considerably slow (i.e., taking several days) for large datasets. Additionally,
we omit algorithms lacking open-source implementation (e.g.,
GreedySPo~\cite{bernstein2020ordering}). Since we focus on linear Gaussian data,
we omit the algorithms that do not comply with this assumption
(e.g.,~\cite{salehkaleybar2020learning,wang2023causal}) in the main comparison.
In \S~\ref{subsec:extension}, we explore the effectiveness of \spot on other
settings.

\parh{Running Time.}~Many algorithms can take a long time to run on large
datasets and the convergence condition is not always satisfied. To make the
comparison fair, we set a timeout of 24 CPU hours for each algorithm. For
algorithms that take more than 24 CPU hours, we halt the algorithm and use the
best-so-far graph as their output. 

\parh{Metrics.}~Under linear Gaussian SCM~\cite{spirtes2000causation}, we can
only up to identify a Markov equivalence class. To make all baselines
comparable, for those algorithms that output an MAG, we convert it to a PAG
(Partial Ancestral Graph) and compare it with the true PAG.
Following~\cite{bhattacharya2021differentiable}, we report the F1 score, TPR
(True Positive Rate), and FDR (False Discovery Rate) on the skeleton,
arrowheads, and tails.

\subsection{End-to-end Comparison}
\label{subsec:rq1}

We report the results in \T~\ref{tab:overall-cmp} on ten datasets with nodes
sampled between $[50, 100]$. We observe that the large datasets pose a
considerable challenge to all methods while \spot\ accurately identifies the
underlying causal structure. ICD fails on one dataset; M3HC and GPS frequently
terminate with convergence warnings on these datasets. We also observe that
constraint-based methods are usually more powerful at identifying the correct
skeleton while all methods have notable difficulties to accurately identify the
arrowheads and tails. In contrast, \spot\ consistently manifests its superiority
across nearly all criteria and yields more precise and stable estimation of
skeletons, arrowheads, and tails. Over the ten datasets, it significantly
improves the performance of its counterpart, ABIC (with p-value of 0.008). We
also investigate the criterion (i.e., Skeleton's FDR) on which \spot\ is
sub-optimal while M3HC is the best. Having said that, we observe that M3HC also
suffers from the lowest TPR and F1 scores on skeleton. Hence, we conclude that
M3HC may be too strict on confirming an edge while \spot\ is more robust and
attains the best F1 scores.

It is worth noting that \spot's strong performance can be attributed to its
ability to effectively balance precision and recall in estimating the underlying
causal relationships. Additionally, the more consistent performance of \spot\
could potentially make it more suitable for various real-world applications
where accuracy and robustness are crucial factors. 

\begin{figure}[h]
	
	\centering
	\includegraphics[width=0.85\linewidth]{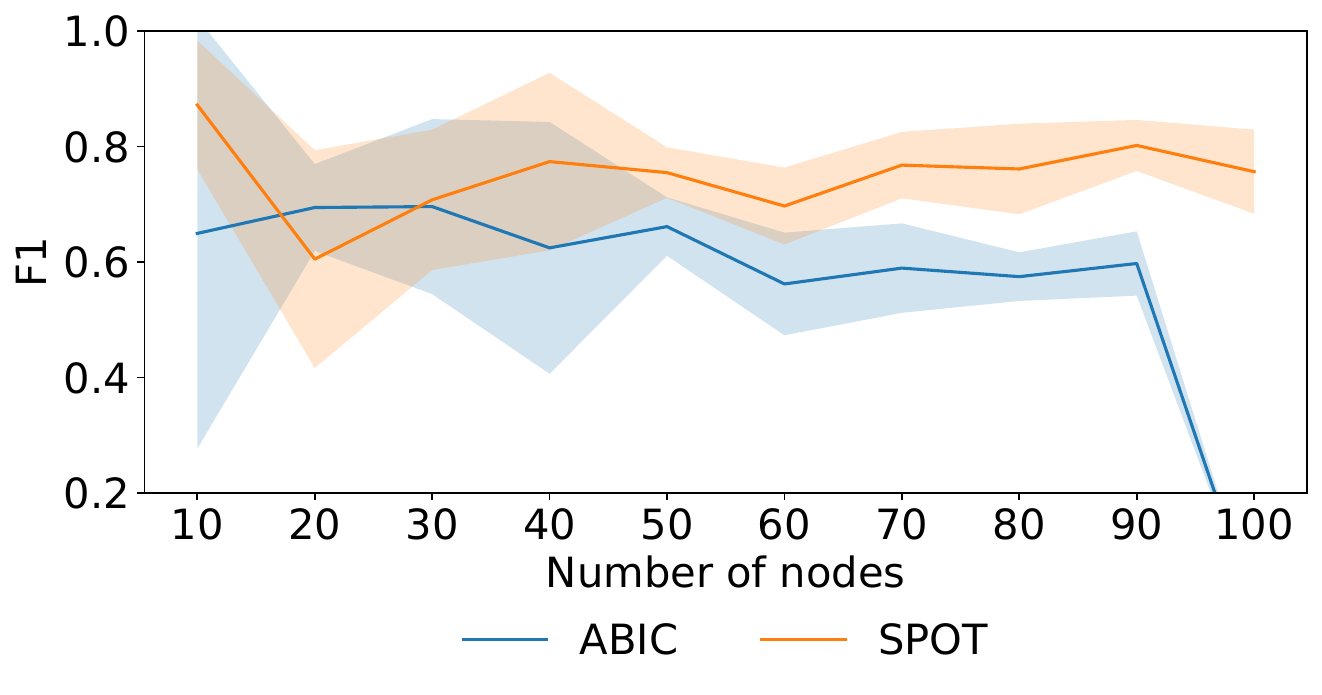}
	
	\caption{Comparison of ABIC and \spot on different node sizes.}
	
	\label{fig:cmp-by-nodes}
\end{figure}

\parh{Comparison by Node Size.}~In addition to the aggregated results, we also
report the edge-wise F1 scores under different node sizes in
\F~\ref{fig:cmp-by-nodes} where the data point of each node size is the average
of three datasets and the error bar is the standard deviation. First, we observe
that \spot\ consistently outperforms all baselines across all node sizes except
for 20-node graphs. For 20-node graphs, potentially due to randomness, though
ABIC slightly outperforms \spot, the difference is not statistically significant
($p=0.51$). In other settings, especially for large graphs, \spot outperforms
ABIC (the second-best method) by a large margin (e.g., $p=0.001$ on 90-node
graphs). Second, the performance of ABIC significantly drops on 100-node graphs
where it fails to learn a meaningful causal structure. In contrast, \spot, by
incorporating the skeleton posterior in the optimization, manifests stable and
strong performance across all node sizes with a small standard deviation.

\subsection{Effectiveness of Skeleton Posterior Inference}

Through this experiment, our goal is to determine if \spot\ is a reliable
skeleton posterior estimator. We assess the posterior quality by calculating the
KL-divergence between the estimated posterior and ground-truth skeleton.
Following earlier research~\cite{lorch2022amortized}, we report AUPRC and AUROC
for edge predictions to evaluate the quality of the estimated posterior.

Using the same datasets from \S~\ref{subsec:rq1}, we compare \spot\ to three
baselines: AVICI~\cite{lorch2022amortized}, FCI* (Nonparametric Bootstrap FCI),
and RFCI* (Nonparametric Bootstrap RFCI). We tried to include more baselines
(e.g., DiBS~\cite{lorch2021dibs} and N-ADMG~\cite{ashman2022causal}) but they
either frequently crash or encounter out-of-memory issues when processing large
graphs used in the experiment. These baselines are not designed for skeleton
posterior inference; however, we sum the probabilities of all edge types,
calibrate to a maximum of 1, and recast their output as a skeleton posterior for
a meaningful comparison in \T~\ref{tab:skl-cmp}. In addition, we also conduct an
ablation study by replacing the ML4S-like model with an end-to-end
Transformer-based model (denoted as \spot (w/ \citet{ke2022learning})) to
justify our design consideration in \S~\ref{subsec:espi}.

\begin{table}[t]
	\caption{Evaluation on skeleton posterior inference (avg.~on ten
	datasets). AUROC and AUPRC: higher is better; KL: lower is better.}
	\label{tab:skl-cmp}
	\centering
	
	\resizebox{0.65\linewidth}{!}{
		\begin{tabular}{l|c|c|c|c}\toprule
			Metric & AVICI & FCI* & RFCI* & \spot \\ \midrule
			AUROC & 0.96 & 0.97 & 0.95 & {\bf 0.99}\\ \midrule
			AUPRC & 0.83 & 0.92 & 0.91 & {\bf 0.97}\\ \midrule
			KL    & 0.05 & 0.06 & 0.10 & {\bf 0.03}\\ \bottomrule
		\end{tabular}
	}
	
\end{table} 
\parh{Comparison with Baselines.}~Overall, we observe that \spot\ consistently
outperforms the baselines across all criteria in \T~\ref{tab:skl-cmp}. This
finding is interpreted as encouraging and reasonable. First, \spot\ employs a
machine learning model to estimate the adjacency probability while FCI and RFCI
apply hard thresholds over p-values to reject edges. Since the relationship
between p-values and adjacency probabilities may be intricate, \spot\ is more
adaptable in capturing these dependencies. Second, despite having a similar
focus on variational inference, AVICI is not intended for use with causally
insufficient data, and the size of the graph may make it more challenging to
perform whole-graph inference. In contrast, \spot\ efficiently resolves this
problem by edge decomposition.

\parh{Ablation Study.}~We also observe that the Transformer-based model (\spot
(w/ \citet{ke2022learning})) yields a lower AUROC (0.95) and AUPRC (0.77) as
well as a higher KL (0.07) than the ML4S-like cascade model (i.e., worse
performance on all criteria). This finding is consistent with our design
consideration in \S~\ref{subsec:espi} that the cascade model is more suitable
for the task of skeleton posterior inference. We presume that the cascade model
effectively decompose the whole graph into edge predictions and reduce the
complexity of the task, which is crucial for large graphs.

\parh{Out-of-Distribution Setting.}~In addition, we also explore the
effectiveness of the domain adaption procedure in \spot\ for out-of-distribution
datasets. To do so, we generate ten ADMGs from the Scale-Free (SF) model as test
datasets and evaluate the enhancement of our domain adaption strategy. On the SF
graphs, the original model trained on ER graphs suffers from a downgrade on KL
(from 0.03 to 0.06). When augmented with the lightweight domain adaption step,
the KL is further reduced to 0.05 (16.7\% enhancement).

\subsection{Extension to Other Settings}
\label{subsec:extension}

Though the primary focus of \spot lies in linear Gaussian data with latent
confounders, we also explore its effectiveness on other types of data. In
particular, we integrate \spot with neural variational ADMG
learning~\cite{ashman2022causal} (N-ADMG) which handles non-linear data and
neural variational DAG learning~\cite{geffner2022deep} (DECI) and
GFlowNets~\cite{deleu2022bayesian} on linear Gaussian data without latent
confounders.

\begin{table}[h]
	
	\caption{Integration with neural variational ADMG learning (avg. on 
	100 sampled ADMGs).}
	\label{tab:admg-cmp}
	
	\centering
	\resizebox{0.65\linewidth}{!}{
		\begin{tabular}{l|c|c|c}\toprule
			 & SHD & FDR & TPR \\ \midrule
			N-ADMG~\cite{ashman2022causal} & 2.61 & 0.39 & 1.0 \\ \midrule
			N-ADMG + \spot & 2.07 \improve{-21} & 0.34 \improve{-13} & 1.0 \\ \bottomrule
		\end{tabular}
	}
	
\end{table}

\parh{Neural Variational ADMG Learning.}~Due to the scalability issue of N-ADMG,
we are unable to apply it to datasets with similar sizes as in
\S~\ref{subsec:rq1}. Instead, we use the same dataset in its original
paper~\cite{ashman2022causal} with five nodes and 2000 samples under the
non-linear Gaussian setting. Here, we use the kernel-based conditional
independence test~\cite{zhang2011kernel} to compute the required statistics in
\S~\ref{subsec:espi}. However, it is worth noting that \spot itself is agnostic
to the functional form of the SCM as long as the distribution is faithful and
the required statistics can be appropriately computed. Since N-ADMG yields a
distribution over ADMGs, we sample 100 ADMGs from the posterior and report the
average performance in \T~\ref{tab:admg-cmp}. Note that, in this setting,
because the ADMG is identifiable, we compare the generated ADMG with the
ground-truth ADMG. We observe that \spot\ significantly improves the performance
of N-ADMG on all criteria. In particular, it reduces the FDR by 13\% and the SHD
by 21\%. In the dataset, N-ADMG is capable of identifying all true edges (i.e.,
TPR=1.0). However, it also introduces many false positives. \spot\ effectively
reduces the number of false positives and improves the overall performance. In
this regard, \spot offers a principled way to discourage potentially superfluous
edges and thus alleviates the overfitting issue.

\begin{table}[h]
	\caption{Integration with neural variational DAG learning (avg. on 100
	sampled DAGs).}
	
	\label{tab:dag-cmp}
	\centering
	\resizebox{0.75\linewidth}{!}{
		\begin{tabular}{l|c|c|c}\toprule
			 & Skeleton F1 & Orientation F1 & SHD \\ \midrule
			DECI~\cite{geffner2022deep} & 0.55 & 0.44 & 121.2 \\ \midrule
			DECI + \spot & 0.58 \improve{+5} & 0.48 \improve{+9} & 118.7 \improve{-2} \\
			\bottomrule
		\end{tabular}
	}
	
\end{table}

\parh{Neural Variational DAG Learning.}~Since MAG is a generalization of DAG,
our framework is naturally applicable to infer the skeleton posterior of DAGs
and thus can be integrated with neural variational DAG learning. To demonstrate
this potential extension of \spot, we integrate it with
DECI~\cite{geffner2022deep} and report the result in \T~\ref{tab:dag-cmp}. Here,
even though \spot is originally designed for MAGs, we can observe a non-trivial
performance boost on DAGs. In particular, \spot improves the F1 score of
skeleton and orientation by 5\% and 9\%, respectively, and reduces the SHD by
2\%. This finding indicates a potential avenue for future research to further
improve the performance of differentiable DAG learning algorithms with skeleton
posterior. In addition to DECI, we also tried to integrate \spot with
GFlowNets~\cite{deleu2022bayesian} which estimates the posterior of DAGs using
generative flow networks. With the same setting, we observe that \spot reduces
the expected SHD of the posterior from 339.8 to 224.4, indicating a 34\%
improvement. We present the full results in the appendix due to the space limit.
\section{Related Work}

\parh{Causal Discovery with Latent Confounders.}~Causal discovery with latent
confounders involves constraint-based methods~\cite{spirtes2000causation,
zhang2008completeness,colombo2012learning,rohekar2021iterative,ma2023xinsight}
and score-based methods~\cite{tsirlis2018scoring, chen2021integer,
claassen2022greedy}. Constraint-based methods establish causal graphs based on
conditional independence~\cite{ma2024enabling}, while score-based methods find
maximal likelihood estimations. Differentiable causal discovery utilizes
continuous optimization techniques~\cite{bhattacharya2021differentiable}. With
the prosperity of its practical applications~\cite{ji2023perfce,
ji2023causality, ji2023benchmarking, ji2023cc, wang2023causal, ma2022noleaks},
\spot advances scalable and accurate causal discovery with latent confounders
for real-world use.

\parh{Posterior Inference of Causal Structure.}~Traditional causal discovery
provides a maximum-likelihood point estimation, but reliability is often
criticized for small sample sizes~\cite{cundy2021bcd}. Providing a
\textit{posterior distribution} is more desired using
MCMC~\cite{viinikka2020towards}, variational
inference~\cite{cundy2021bcd,lorch2021dibs,lorch2022amortized}, reinforcement
learning~\cite{yang2022reinforcement,agrawal2018minimal}, generative flow
network~\cite{deleu2022bayesian}, or supervised learning~\cite{ma2022ml4s}. All
methods estimate DAGs' posterior distributions and is infeasible for MAGs. Among
these methods, AVICI~\cite{lorch2022amortized} is the most relevant to \spot as
it also uses amortized variational inference. In a nutshell, it trains an
inference model over samples from a known (static) simulator and learn the DAGs
from data. In the context of ADMG learning, N-ADMG~\cite{ashman2022causal}
presents a variational inference method to estimate the posterior of ADMGs on
non-linear data. However, it is not scalable well to large datasets, as shown in
\S~\ref{sec:eval}. \spot trains an inference model on these samples from
static/dynamic simulators and alleviate the additional assumption of the data. 

\parh{The Role of Skeleton in Causal Discovery.}~Skeleton learning is crucial
for constraint-based methods~\cite{yu2016review} and affects overall
performance. Score-based DAG learning methods like MMHC use skeletons to reduce
search space~\cite{tsamardinos2006max}. Ma et al.~improve the performance of
NOTEARS via a more precise skeleton learned by ML4S~\cite{ma2022ml4s}. \spot
demonstrates the importance of skeleton learning in differentiable causal
discovery. Taking these evidences into consideration, it may become clear that
skeleton learning can serve as the backbone for general causal discovery tasks.

\section{Conclusion}

In this paper, we introduce a framework for differentiable causal discovery in
the presence of latent confounders with skeleton posterior. To this end, we
propose \spot, which features a highly efficient variational inference algorithm
to estimate the underlying skeleton distribution from given observational data.
It also provides a stochastic optimization procedure to seamlessly incorporate
the skeleton posterior with the differentiable causal discovery pipeline in an
``out-of-the-box'' manner. The results of our experiments are highly encouraging
and show that \spot\ outperforms all existing methods to a notable extent.

\begin{acks}
The authors would like to thank the anonymous reviewers for their valuable
comments. The authors from HKUST were supported in part by a RGC CRF grant under
the contract C6015-23G.
\end{acks}

\clearpage

\setcounter{section}{0}
\renewcommand{\thesection}{\Alph{section}}

\section{Preliminary}

To keep the paper self-contained, we provide additional definitions and
notations used in the main text.

\begin{definition}[Directed Cycle]
    Given an ADMG $G$ with a set of nodes $V$ and a set of edges $E$, a directed
    cycle exists when there is a directed path from a node $V_i$ back to itself.
\end{definition}

\begin{definition}[Almost Directed Cycle]
    Given an ADMG $G$ with a set of nodes $V$ and a set of edges $E$, an almost
    directed cycle exists when there is a bidirected edge $V_i \leftrightarrow
    V_j$ such that $V_i \in \bm{An}_G(V_j)$. $\bm{An}_G(V_j)$ denotes the set of
    ancestors (by directed paths) of $V_j$ in $G$.
\end{definition}

\begin{definition}[Bow]
	Given an ADMG $G$ with a set of nodes $V$ and a set of edges $E$, a bow
	exists when there are two edges $V_i \rightarrow V_j$ and $V_j
	\leftrightarrow V_j$.
	
\end{definition}

\begin{definition}[Bow-free ADMG]
	An ADMG is bow-free if it does not contain any directed or a bow.
\end{definition}

A path $(X,W_1,\cdots,W_k,Y)$ is said to be \textit{blocked} by $Z\subseteq
\bm{X}\setminus\{X,Y\}$ if there exists a node $W_i\in \{W_1,\cdots,W_k\}$ such
that a) $W_i$ is not a collider but a member of $Z$, or b) $W_i$ is a collider
but not an ancestor of any nodes of $Z$. We now introduce \textit{m-separation}.

\begin{definition}[m-separation~\cite{zhang2008completeness}]
  \label{def:mseparation}
  $X,Y$ are m-separated by $Z$ (denoted by $X\sep Y\mid Z$) if all paths
  between $X,Y$ are blocked by $Z$. 
\end{definition}

\section{Implementation of SCL Models in \spot}

In accordance with ML4S~\cite{ma2022ml4s}, we implement the supervised causal
learning (SCL) model in \spot\ as a series of cascade
xgboost~\cite{chen2016xgboost} classifiers with default hyperparameters. For
linear datasets, we use Fisher-z test to calculate the conditional independence
and constitute the features for the classifier. For non-linear datasets, we use
the kernel-based conditional independence test~\cite{zhang2011kernel} instead.
We use the output of the last layer of the cascade xgboost classifiers as the
posterior probability of the presence of an edge.

\begin{figure}[h]
	\centering
	\includegraphics[width=\linewidth]{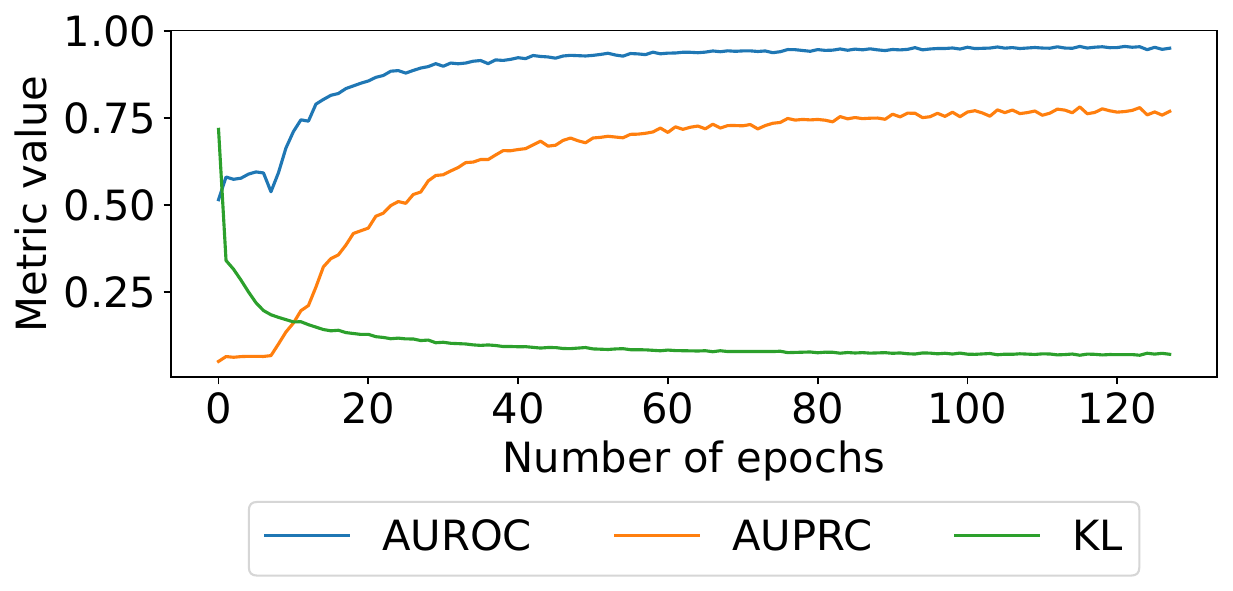}
	\caption{Convergence of the model in \citet{ke2022learning}.}
	\label{fig:ke2022-convergence}
\end{figure}

Since the model used in \citet{ke2022learning} is not publicly available, we
made our best effort to replicate the model based on the description in the
paper. The experiments are conducted on a server with NVIDIA A6000 GPU and 256GB
RAM. The model is trained with 128 epochs and a batch size of 1. We use the Adam
optimizer with a learning rate of 0.0003. When trained with 128 epochs, the
metrics is seen to converge, as shown in \F~\ref{fig:ke2022-convergence}. On the
large-scale causal graphs with 50--100 nodes, each dataset contains 1,000
observational samples. However, the GPU (with 48GB memory which is the largest
GPU memory available to us) can at most handle 600 samples at a time. Hence, we
drop the last 400 samples in each dataset. This issue, to a certain extent,
indicates the scalability limitation of end-to-end supervised causal learning
models and promotes the necessity of ML4S-like methods, which are much more
scalable and efficient.

\section{Ablation Study on Sparsity Prior}

In our optimization procedure, we enforce a heuristic in which sparsifying
proposals are always accepted. This design attempts to analogize a range of the
regularization term in score-based causal discovery algorithms where sparse
causal graphs are preferred over dense ones. To evaluate the impact of this
design, we conduct an ablation study by removing the sparsity prior from the
optimization procedure. The results are shown in \T~\ref{tab:ablation-sparsity}.

\begin{table}[h]
	\caption{Ablation study on sparsity prior.}
	\label{tab:ablation-sparsity}
	\centering
	\resizebox{0.65\linewidth}{!}{
		\begin{tabular}{l|c|c|c}\toprule
			 & Skeleton F1 & Head F1 & Tail F1 \\ \midrule
			ABIC & 0.84 & 0.76 & 0.67 \\ \midrule
			\spot w/o Sparsity & 0.90 & 0.81 & 0.71 \\
			\spot & 0.91 & 0.86 & 0.78 \\
			\bottomrule
		\end{tabular}
	}
\end{table}

Overall, we observed that the strategy indeed provides a non-trivial improvement
to performance and believe that such sparsity consideration plays an important
role in ``pruning'' dense causal graphs.

\section{Compute Time}

We refrain from providing a table regarding compute time because different
algorithms are computed on different architectures, making it difficult to draw
meaningful conclusions. For instance, FCI can only use a single core, whereas
ABIC/SPOT can utilize up to 32 cores, and N-ADMG primarily relies on GPU.
Additionally, on some large datasets, ABIC may fail to terminate even after
running for one week. These issues complicate direct comparisons. Following your
suggestion and consistent with the main setup in our paper, we report the
compute times for ABIC and SPOT on the same computing architecture below.

\begin{table}
	\caption{Average compute time (in seconds) for ABIC and \spot.}
	\label{tab:compute-time}
	\centering
	\begin{tabular}{l|c}\toprule
		Algorithm & Avg. Time (sec.) \\ \midrule
		ABIC & 3139 \\
		\spot & 2883 \\
		\bottomrule
	\end{tabular}
\end{table}

Overall, we believe SPOT enables lower compute time as well as better
performance.

\section{Comparison by Different Number of Nodes}

\begin{figure}[h]
	\centering
	\includegraphics[width=0.85\linewidth]{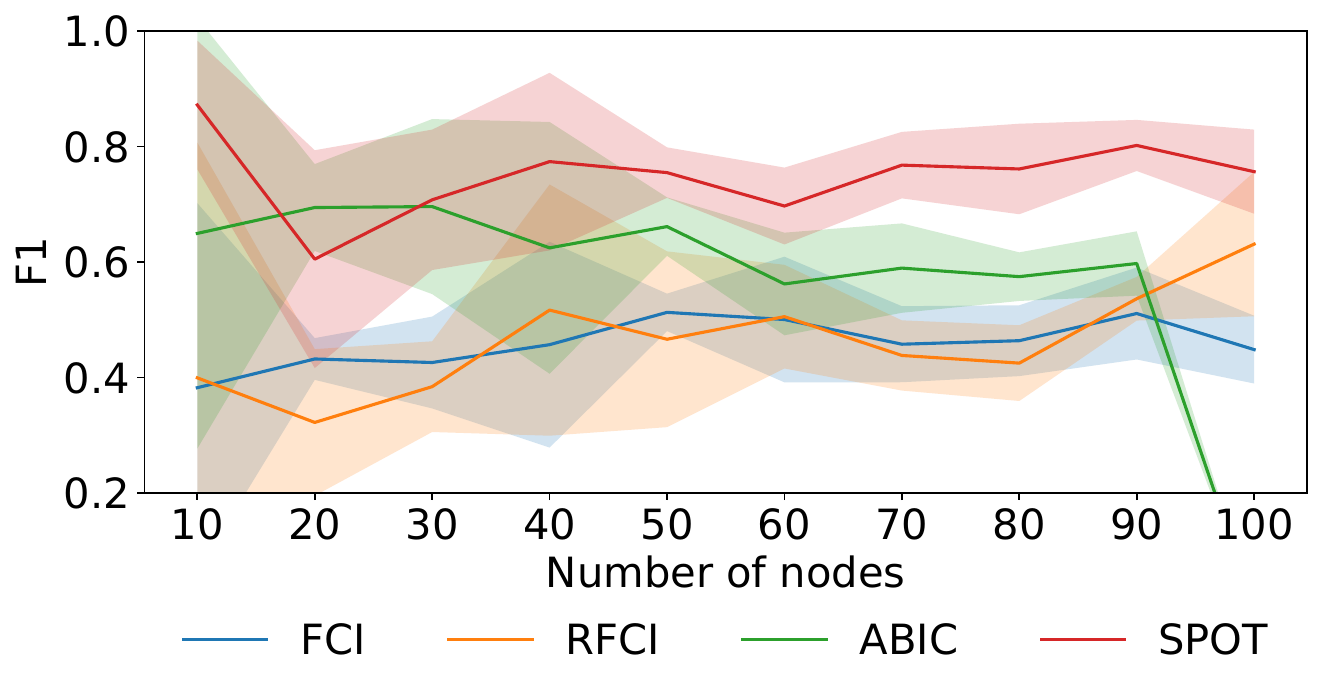}
	\caption{Comparison of methods on different node sizes.}
	\label{fig:cmp-by-nodes-full}
\end{figure}

We report the full results of the comparison of different methods on different
methods, including FCI, RFCI, ABIC and \spot. Aligned with our observations in
\S~\ref{subsec:rq1}, \spot manifests superior performance in all node sizes 
except on 20-node datasets.

\section{Integration with GFlowNets}

\begin{table}[h]
	\caption{Integration with GFlowNets~\cite{deleu2022bayesian}.}
	\label{tab:gflownets-cmp}
	\centering
	\resizebox{0.65\linewidth}{!}{
		\begin{tabular}{l|c|c|c}\toprule
			 & F1 & Precision & Recall \\ \midrule
			GFlowNets~\cite{deleu2022bayesian} & 0.22 & 0.14 & 0.45 \\ \midrule
			GFlowNets + \spot & 0.25 & 0.18 & 0.44 \\
			\bottomrule
		\end{tabular}
	}
\end{table}

We also explore the integration of \spot with GFlowNets~\cite{deleu2022bayesian}
to further improve the performance of differentiable causal discovery. GFlowNets
is a generative flow network that learns the posterior distribution of DAGs. We
use the same experimental setup as in \S~\ref{subsec:extension}, and the results
are shown in \T~\ref{tab:gflownets-cmp}. We observe that \spot\ improves the
performance of GFlowNets by 13.6\% in F1 score and 28.6\% in precision at the
cost of a slight decrease in recall. This result demonstrates the synergistical
effect of \spot in differentiable causal discovery.

\section{Real-world Data}

\begin{figure}[!ht]
	\centering
	
	\includegraphics[width=0.75\columnwidth]{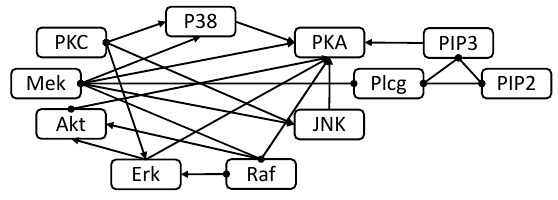}
	\caption{Application on Sachs dataset.}\label{fig:sachs}
	
\end{figure}
We also explore the application of our proposed framework on
Sachs~\cite{sachs2005causal}, a real-world dataset with 853 entries on protein
expressions involved in human immune system cells. \F~\ref{fig:sachs} presents
the promising results of \spot\ applied to the Sachs dataset. Since every edge
endpoint of Sachs is unidentifiable, its PAG corresponds exactly to the
skeleton. Therefore, we evaluate the learned graph's quality at the skeleton
level. The discovered skeleton in \F\ref{fig:sachs} achieves an F1 score of
0.63. Although there is a decrease in performance compared to the results in
\S~\ref{subsec:rq1}, it still significantly outperforms ABIC (with an F1 score
of 0.48) and provides satisfactory results. The decrease in performance can be
partially attributed to the non-Gaussian nature of real-world data, which
affects the fundamental premise of ABIC, the backbone of our framework. Since
\spot\ itself (i.e., skeleton posterior inference phase) is agnostic to
Gaussianity, it delivers an impressive 31.2\% improvement over vanilla ABIC. We
believe that the performance of \spot\ has the potential for even further
improvement by incorporating differentiable methods that handle non-Gaussian
data in future work.

\bibliographystyle{ACM-Reference-Format}
\bibliography{main}

\end{document}